# A Low-Cost Lane-Following Algorithm for Cyber-Physical Robots

Gupta Archit
School of Computer Science and Engineering, Nanyang Technnological University

Assoc Prof Arvind Easwaran
School of Computer Science and Engineering, Nanyang Technnological University

***Abstract –*** Duckiebots are low-cost mobile robots that are widely used in the fields of research and education. Although there are existing self-driving algorithms for the Duckietown platform, they are either too complex or perform too poorly to navigate a multi-lane track. Moreover, it is essential to give memory and computational resources to a Duckiebot so it can perform additional tasks such as out-of-distribution input detection. In order to satisfy these constraints, we built a low-cost autonomous driving algorithm capable of driving on a two-lane track. The algorithm uses traditional computer vision techniques to identify the central lane on the track and obtain the relevant steering angle. The steering is then controlled by a PID controller that smoothens the movement of the Duckiebot. The performance of the algorithm was compared to that of the NeurIPS 2018 AI Driving Olympics (AIDO) finalists, and it outperformed all but one finalists. The two main contributions of our algorithm are its low computational requirements and very quick set-up, with ongoing efforts to make it more reliable.

**Keywords –** autonomous driving, computer vision, robotics, lane following, duckietown

## 1 INTRODUCTION

Duckietown are low-cost mobile robots widely used for research and education in the fields of robotics, computer vision, and machine learning. Although there exist self-driving algorithms for the Duckietown platform, they are either too complex—occupying large parts of memory—or take too long to train. As a result, there are no existing algorithms for the Duckietown system that perform low-computation lane-following, thus stalling progress in fields building on top of autonomous driving. One example of such an application is out-of-distribution detection [1].

There has been prior work in the field of end-to-end lane-detection—especially with respect to Duckietown. Most work pertaining to the Duckietown ecosystem pertains with either reinforcement learning or transfer learning. The most common technique used for transfer learning in the self-driving domain is domain randomization, and has been combined with reinforcement learning to solve lane following [2] - [4]. Other end-to-end lane following methods have also been proposed for the same [5], [6] but they all use deep learning. The main drawback with these methods are the extensive set-up and long training times which makes them inflexible and tedious for applications aimed towards building and testing algorithms building on top of self-driving.

As part of the NeurIPS 2018 AI Driving Olympics competition, several groups presented self-driving algorithms for the Duckiebot [7], [8]. However, given the rules of the competition, computation was done on a separate and more-powerful machine. Moreover, these algorithms lack detailed implementations. As a result, it is not possible to test the spatial-temporal efficiency of safety monitoring algorithms, making this suite of algorithms unsuitable for the same reasons as above. Although more recent versions of the AI Driving Olympics have taken place, there exists a dearth of results or implementations.

While the algorithm proposed in this paper does on-device computation, it is useful to briefly discuss relevant algorithms presented at the competition. Participant Jon Plante built a computer vision-based algorithm that pre-processed the input image, and coupled it with a case-based controller and inverse kinematics to produce motor signals. This approach is similar to the approach presented in this paper, with the core difference being on-device computation. Another participant Vincent Mai fine-tuned a ROS-based lane-following baseline and improved its performance. Other approaches explored by finalists were primarily deep reinforcement learning based.

In summary, most approaches to end-to-end lane detection either cannot be run efficiently on a Duckiebot due to computational constraints, or require extensive set-up and training—making them unsuitable for off-the-shelf and easy implementation on Duckiebots. However, the approaches discussed above are by no means exhaustive. In this paper, the algorithm developed employs classical computer vision methods to detect the central lane on a multi-lane track, and

**Algorithm 1** Lane-Following Algorithm

```
procedure IMAGE PROCESSING
    Receive frame y and crop the top half, and convert it to greyscale
    Apply thresholding and masking to obtain the central lane marking
    Apply Gaussian blur, Canny edge detection, & Hough line transformation
    Convert the obtained line segments to straight lines
    Calculate the angle of deviation from the obtained straight lines using trigonometry
    Use the calculated angle to update position information
    Return calculated angle
end procedure

procedure CONTROL LOGIC
    Analyse the positional information to predict direction
    Convert the direction into a steering angle using the calculated angle
    Return steering angle
end procedure
```

*Figure 1: Algorithm Pseudocode*

then follows it. The algorithm was deployed on a DB21M Duckiebot equipped with an NVIDIA Jetson Nano 2GB.

This paper is divided into 5 sections—Introduction, Duckietown Framework, Algorithm Design, Experimental Results, and Conclusion & Further Work.

## 2 DUCKIETOWN FRAMEWORK

Duckietown is an open, inexpensive, and flexible platform for robotics education and research. The platform constitutes small autonomous vehicles called Duckiebots built from off-the-shelf components, and cities called Duckietowns complete with infrastructure such as roads, signage, and obstacles [9]. The Duckietown platform also includes certain specifications for maps and a simulation environment, but we do not exploit these in the scope of this paper.

The Duckiebot used in this paper is the DB21M with all computation taking place on a NVIDIA Jetson Nano 2GB. The sensor suite includes a camera, a time of flight sensor, an inertial measurement unit, and wheel encoders.

More recent versions of the Duckiebot (DB21) feature an expanded onboard memory from 32GB to 64GB and a tweaked chassis design for reduced complexity and increased stiffness. The software architecture of the Duckietown platform is designed for fast and modular development. The several robotics ecosystem functionalities offered on the platform are encapsulated in Docker containers—including ROS (Robotic Operating System) and Python code. Docker ensures reproducibility of code and eases modularity, which makes it easy to substitute or add individual functional blocks without running into compatibility issues. ROS controls communication between functionalities such as perception, planning, and high- and low-level control, thereby providing abstraction for the hardware later. Moreover, the Duckietown platform adds another layer of abstraction using a homemade operating system (duckie-OS) which handles the ROS nodes running inside Docker containers. [10]

## 3 ALGORITHM DESIGN

Our algorithm design is three-fold, comprising image processing, control logic, and a PID controller. The image processing module takes in raw images from the Duckiebot and outputs the central angle amounting to the angle of deviation between the Duckiebot and the central lane. It also updates some global variables that are used to maintain the vehicle's internal state. The PID controller receives a steering angle and configures the motor speeds for each of the two wheels to ensure smooth traversal of the track. It has been borrowed from [1]. The algorithm pseudocode has been provided in figure 1.

### 3.1 IMAGE PROCESSING

The image processing module takes incoming frames, pre-processes them using computer vision techniques, and then uses trigonometry to determine the angle of deviation of the vehicle from the central line.

The incoming image is pre-processed using image cropping, colour conversion, and thresholding. Once pre-processed, the central lane marking is identified using Gaussian blur, Canny edge detection, and a Hough line transformation. The line segments obtained are used to create two lines indicating the direction the vehicle should move in.

From these lines, an angle is obtained using trigonometry that guides the direction of the vehicle.

## 3.2 CONTROL LOGIC

The control logic interacts with the various functions to streamline image processing and angle calculation. It combines the angle calculated with past information to output a steering angle. The past information is stored as a single integer indicating what direction the vehicle took previously.

## 4 EXPERIMENTAL RESULTS

In order to quantify the performance of our algorithm, we compared our algorithm's performance with those of the finalists of the 2018 NeurIPS AI Driving Olympics (AIDO) competition [8]. We recreated the track used in the AIDO Lane Following (LF) event as closely as possible. The original track comprised 18 individual blocks and the vehicles traversed the track 5 times—starting from a different block each time. Two performance metrics were used—time travelled before the vehicle exited the lane (in seconds) and the number of tiles traversed. The final performance of the algorithm was then calculated as the sum of these two metrics across all 5 runs.

| Contestant | Round 1 | Round 2 | Round 3 | Round 4 | Round 5 | Cumulative |
|---|---|---|---|---|---|---|
| Wei Gao | 3.3/1 | 3.9/1 | 2.0/1 | 23.0/12 | 5.0/3 | 37/18 |
| SAIC Moscow | 6.0/3 | 2.0/1 | 2.0/1 | 3.0/1 | 2.0/1 | 15/7 |
| Team JetBrains | 16.0/2 | 1.0/0 | 4.0/1 | 0.0/0 | 8.0/1 | 29/4 |
| Jon Plante | 18.0/2 | 1.0/0 | 7.0/3 | 3.0/1 | 5.0/3 | 34/9 |
| Vincent Mai | 2.0/1 | 1.0/0 | 3.0/2 | 14.0/1 | 3.0/2 | 23/9 |
| Our Algorithm | 12.8/4 | 7.9/2 | 7.1/2 | 4.8/2 | 4.8/2 | **37.4/12** |

*Figure 2: Results*

In order to simulate these conditions, we chose 5 unique starting blocks—each roughly 3 blocks apart. We then ran our Duckiebot 5 times in succession—one run per starting block. The codebase was built once before the first run started, and no further changes were made to it.

As shown in figure 2, our algorithm outperformed all models but one. However, our algorithm has one drawback which makes it unreliable—sharp turn-taking. While the angles calculated are accurate, the vehicle often responds a little too late. This is seen most often as the vehicle takes turns after it goes off path. One possible reason could be late processing of the most recent frame as the algorithm is still computing earlier frames.

## 5 CONCLUSION & FURTHER WORK

The algorithm presented in this paper provides a cheap and quick method to bring lane-following to cyber-physical systems such as Duckiebots. Despite struggling with turn-taking, our algorithm outperformed all but one from NeurIPS 2018 AI Driving Olympics.

The algorithm presented has scope of improvement. The control logic can be tweaked to detect turns earlier and respond accordingly. As a whole, the algorithm presented in this paper could be expanded upon by incorporating data from the edge lane markings to better estimate the position of the vehicle and inform its next move. Moreover, the current algorithm can be used for data collection to create a deep learning model to abstract the control logic away and build a more robust algorithm.

## ACKNOWLEDGMENT

I would like to acknowledge the funding and support from Nanyang Technological University–URECA Undergraduate Research Program for this research project. I also extend my heartfelt gratitude to Michael Yuhas for his extensive guidance and support. I would also like to express gratitude to all the researchers and hobbyists whose work we have built upon.